\documentclass[a4paper,10pt]{article} 

\usepackage[T1]{fontenc}
\usepackage[utf8]{inputenc}
\usepackage{lmodern}

\usepackage{color}
\usepackage{longtable}

\usepackage[cmex10]{amsmath}
\usepackage{amssymb,amsfonts}
\usepackage{wasysym}

\usepackage{graphicx}
\usepackage{algorithmic}
\usepackage{array}
\usepackage{url}
\usepackage{enumitem}
\usepackage{multicol}
\usepackage[bb=boondox]{mathalfa}

\usepackage{todonotes}
\usepackage{cite}
\usepackage{hyperref}

\usepackage[ruled,vlined]{algorithm2e}

\newcommand{\vect}[1]{\boldsymbol{#1}}
\newcommand{\node}[1]{\mathrm{#1}}

\newcommand{\ds}{\displaystyle}

\newcommand{\Rr}{{\mathbb R}}

\newcommand{\Ll}{\mathbf{L}}
\newcommand{\Aa}{\mathbf{A}}
\newcommand{\Dd}{\mathbf{D}}
\newcommand{\Kk}{\mathbf{K}}

\newcommand{\Exp}{\mathrm{E}}
\newcommand{\Power}{\mathrm{P}}
\newcommand{\WIM}{{W}^{(\mathrm{IM})}}

\newcommand{\1}{\mathrm{1\hspace{-1mm}l}}

\newcommand{\subind}[2]{\raisebox{-0.5pt}[0mm][0mm]{\scalebox{.55}{$#1,\hspace{-1pt} #2$}}}

\DeclareMathOperator{\argmax}{arg\max}

{\bfseries}{\itshape}
{\bfseries}{\itshape}
{\bfseries}{\itshape}
{\bfseries}{\itshape}

\newtheorem{dfn}{Definition}{\bfseries}{\rmfamily}
\newtheorem{rem}{Remark}{\bfseries}{\rmfamily}
\newtheorem{exa}{Example}{\bfseries}{\rmfamily}
\newtheorem{ppt}{Property}{\bfseries}{\rmfamily}

\usepackage{authblk}

\title{Kernel-Based Models for Influence Maximization on Graphs based on Gaussian Process Variance Minimization}

\author[1]{S. Cuomo 
\thanks{\href{mailto:salvatore.cuomo@unina.it}{salvatore.cuomo@unina.it}}}
\affil[1]{University of Napoli}

\author[2]{W. Erb 
\thanks{\href{mailto:wolfgang.erb@unipd.it}{wolfgang.erb@unipd.it}}}
\affil[2]{University of Padova}

\author[3]{G. Santin 
\thanks{\href{mailto:gsantin@fbk.eu}{gsantin@fbk.eu}}}
\affil[3]{Bruno Kessler Foundation}

\date{\today}

\begin{document}
\maketitle

\begin{abstract} 
The inference of novel knowledge, the discovery of hidden patterns, and the uncovering of insights from large amounts of data from a multitude of sources make Data Science (DS) to an art rather than just a mere scientific discipline. The study and design of mathematical models able to analyze information represents a central research topic in DS. In this work, we introduce and investigate a novel  model for influence maximization (IM) on graphs using ideas from kernel-based approximation, Gaussian process regression, and the minimization of a corresponding variance term.
Data-driven approaches can be applied to determine proper kernels for this IM model and machine learning methodologies are adopted to tune the model parameters. Compared to stochastic models in this field that rely on costly Monte-Carlo simulations, our model allows for a simple and cost-efficient update strategy to compute optimal influencing nodes on a graph. In several numerical experiments, we show the properties and benefits of this new model. 

\end{abstract}

\section{Introduction}
Online social platforms such as Facebook, Twitter, LinkedIn, and Tumblr, are pervasively present in our daily life. In these virtual communities, people are emotionally connected and share several kinds of relationships as friendship, common interests, or news. Users in a social network produce a huge amount of digital information. For example, the publishing of 100 posts with photos on Facebook generates already an estimated traffic of 500MB.  
(see \url{https://www.att.com/support/data-calculator} for such estimates). For this reason, Data Science (DS) techniques specifically developed for social network analysis and mining display a remarkable growth and provide promising opportunities for the inference of novel knowledge.

From a mathematical point of view, the natural framework for social network analysis is to represent network users and mutual interactions by a {\em graph}, a conceptual structure where network users correspond to abstractions called {\em nodes} and each related pair of nodes is called {\em link} or {\em edge}. 

Graph Signal Processing (GSP), as for instance outlined in \cite{Ortega2018,StankovicDakovicSejdic2019},
is a research field that provides a general mathematical framework to analyze and process data organized in graph structures. In this work, we will mainly focus on kernel-based methods on graphs. Such methods were recently studied in \cite{erb2019b,erb2020b} in terms of interpolation and approximation of graph signals with graph basis functions (GBFs) and partition of unity methods. In \cite{erb2020}, an application of feature-augmented graph basis functions in a semi-supervised machine learning framework was studied. Further theoretical and computational aspects related to uncertainty principles on graphs can be found in \cite{erb2019}.

In this work, we focus on the analysis of kernel-based models for the problem of Influence Maximization (IM) on graphs. 
There exist several applications such as expert finding, topic monitoring, or disease outbreak detection that involve the resolution of the IM problem. More specifically, IM was addressed in \cite{Domingos} for finding potential customers in a market by adopting a probabilistic method based on Markov Random Fields \cite{Rue}. A natural greedy strategy for the solution of a discrete stochastic optimization problem to study the influence of users in a social network was presented in \cite{Kempe}.

Formally, given a graph $G$ with a set $V$ of nodes, a diffusion model ${\Phi}$, and a positive real number $N$ (called \emph{budget}), the IM problem aims at finding a subset of nodes $W_{N}^{(\mathrm{IM})}$ such that 
\begin{equation}\label{im}
W_{N}^{(\mathrm{IM})}=\mbox{arg}\max_{|W|=N}\Phi(W)
\end{equation}
with $W \subseteq V$. In other words the problem (\ref{im}) 
consists in calculating a set $W_{N}^{(\mathrm{IM})}$ of $N$ nodes such that the influence spread of $W_{N}^{(\mathrm{IM})}$, under the given diffusion model $\Phi$, is maximized.

In (\ref{im}), two crucial computational issues have to be considered: (i) the influence computation, i.e., the computation of the influence spread $\Phi(W)$ from the initial subset of nodes $W$; (ii) the maximization of the influence spread. Both problems might be computationally expensive for large graphs, and it is well known that (\ref{im}) is NP-hard in complexity. Strategies to overcome this issue are generally based on greedy approaches adopted in IM to deal with the combinatorial source, while influence computation is addressed using Monte Carlo simulations. Important for the performance of the greedy algorithm is the choice of the influence function $\Phi$. Good approximation results are guaranteed when the influence function $\Phi$ is monotone and submodular \cite{Nemhauser}. 

In this paper, 
we introduce and discuss a novel model for IM on graphs. After some mathematical considerations on kernel-based approximation, we propose a greedy algorithm where the function $\Phi$ is defined by the variance term of a Gaussian process regression (its square root will be referred to as Power function). This Gaussian process variance depends solely on the covariance kernel of the process. Compared to a stochastic diffusion model in which a stochastic simulation is necessary to calculate the spread, our kernel-based variance term $\Phi$ is deterministic. In the corresponding P-greedy algorithm for the approximate solution of the optimization problem a simple update scheme can be implemented in order to reduce the computational costs. For this reason, one of the main advantages of our deterministic IM model is a low computational complexity. Furthermore, our kernel-based approach offers a large flexibility in the choice of the covariance kernel, allowing to select the kernel according to a data-driven learning approach. 
We will present several numerical experiments to highlight the importance of this approach in DS foundations, and, in particular, for IM on graphs. 

The paper is organized as follows. Section \ref{sec:GP_graphs} recalls the necessary tools from GSP and graph spectral theory. In particular, a connection between positive definite Graph Basis Functions (GBFs) and stationary Gaussian Processes on graphs is established.
Based on these tools, Section \ref{sec:new_model} introduces the new model for influence maximization, which is then approximated via the P-greedy algorithm in Section \ref{sec:p_greedy}.
Section \ref{sec:numerics} provides several numerical experiments for IM on graphs and compares the new method with existing techniques.

\section{Gaussian process regression and kernels on graphs}\label{sec:GP_graphs}

\subsection{Preliminaries on graph theory} \label{sec:graphtheory}

In this work, we will use the following mathematical terminology to describe a graph $G=(V,E,\mathbf{L})$ and its elements:
\begin{enumerate}[label = (\roman*)]
\item The vertex set $V=\{\node{v}_1, \ldots, \node{v}_{n}\}$ describes the $n$ nodes $\node{v}_1, \ldots, \node{v}_{n}$ of the graph $G$.  
\item The set $E \subseteq V \times V$ contains all possible edges
$e_{\subind{i}{i'}} = (\node{v}_i, \node{v}_{i'}) \in E$, $i \neq i'$, of the graph $G$. We assume that the graph is undirected, i.e., that with $e_{\subind{i}{i'}} \in E$ also  $e_{\subind{i'}{i}}$ is contained in $E$ and both describe the same edge. With these assumptions, the graph $G$ is simple, i.e. it contains no multiple edges and no self-loops.  
\item The strength of the connection between the vertices in $V$ is modeled with a symmetric matrix $\Ll \in \Rr^{n \times n}$ usually referred to as graph Laplacian. We assume that $\Ll$ is a general matrix of the form (cf. \cite[Section 13.9]{GodsilRoyle2001})
\begin{equation} \label{eq:generalizedLaplacian}
\ds {\begin{array}{ll}\; \Ll_{\subind{i}{i'}}<0& \text{if $i \neq i'$ and $\node{v}_{i}, \node{v}_{i'}$ are connected, i.e., $e_{\subind{i}{i'}} \in E$}, \\ \; \Ll_{\subind{i}{i'}}=0 & \text{if $i\neq i' $ and $\node{v}_{i}, \node{v}_{i'} $ are not connected}, \\ \; \Ll_{\subind{i}{i}} \in \Rr & \text{for $i \in \{1, \ldots, n\}$}.\end{array}}
\end{equation}
\end{enumerate}

Here, the connection weights of the edges $e_{\subind{i}{i'}} \in E$ are encoded in 
the negative non-diagonal elements $\Ll_{\subind{i}{i'}}$, $i \neq i'$, of $\Ll$.  They are linked to the symmetric adjacency matrix $\Aa$ in the following way:
\begin{equation*}
    \mathbf{A}_{\subind{i}{i'}} := 
  \begin{cases}
    - \Ll_{\subind{i}{i'}}, & \text{if $e_{\subind{i}{i'}} \in E$}, \\
    0, & \text{otherwise}.
  \end{cases}
\end{equation*}
There are several possibilities to fix the diagonal entries of $\Ll$. The most common form is the standard graph Laplacian $\Ll_S$ defined as 
\begin{equation} \label{eq:standardLaplacian} \Ll_S = \mathbf{D} - \Aa\end{equation}
with the diagonal degree matrix $\Dd$ given by 
\begin{equation*}
    \mathbf{D}_{\subind{i}{i'}} := 
  \begin{cases}
    \sum_{j=1}^n \mathbf{A}_{\subind{i}{j}}, & \text{if } i=i', \\
    0, & \text{otherwise}.
  \end{cases}
\end{equation*}
A major advantage of the definition \eqref{eq:standardLaplacian} is that the matrix $\Ll_S$ is positive semi-definite. Related to this is also the normalized graph Laplacian $\Ll_N$ given by 
$$\Ll_N = \Dd^{-1/2} \Ll_S \Dd^{-1/2}.$$
For the normalized Laplacian $\Ll_N$ all diagonal entries are equal to $1$ and the spectrum is contained in the interval $[0,2]$. More detailed information regarding combinatorial graph theory and the properties of the different graph Laplacians can be found in \cite{Chung,GodsilRoyle2001}.

\subsection{The Fourier transform of graph signals}
Information on the vertices of the graph is usually encoded in terms of graph signals $x: V \rightarrow \mathbb{R}$. As the vertices $\node{v}_i$, $i \in \{1, \ldots, n \}$ are ordered, we have a natural representation of a signal as a vector $x = (x(\node{v}_1), \ldots, x(\node{v}_n))^{\intercal}\in \mathbb{R}^n$. The corresponding $n$-dimensional vector space $\mathcal{L}(G)$ of graph signals can therefore be regarded as an Euclidean space with the inner product
$$y^\intercal x := \sum_{i=1}^n x(\node{v}_i) y(\node{v}_i).$$ 
If we define the unit signals $\delta_{\node{v}_{i'}} \in \mathcal{L}$ as $\delta_{\node{v}_{i'}}(\node{v}_i) = \delta_{\subind{i}{i'}}$ for $i,i' \in \{1, \ldots,n\}$, the system $\{\delta_{\node{v}_1}, \ldots, \delta_{\node{v}_n}\}$ forms a canonical orthonormal basis of $\mathcal{L}(G)$. 

The Fourier transform on graphs is a natural extension of related Fourier concepts in classical harmonic analysis and provides an indispensable tool for the analysis and processing of graph signals. It is defined by the eigendecomposition of the graph Laplacian (the most common choices for the graph Fourier transform are $\Ll_S$ or $\Ll_N$)
\begin{equation*}
\mathbf{L}=\mathbf{U}\mathbf{M}_{\lambda} \mathbf{U^\intercal}.
\end{equation*}
where $\mathbf{M}_{\lambda} = \mathrm{diag}(\lambda) = \text{diag}(\lambda_1,\ldots,\lambda_{n})$ 
is the diagonal matrix containing the increasingly ordered eigenvalues $\lambda_i$, $i \in \{1, \ldots, n\}$, of $\mathbf{L}$ as diagonal entries, and $\mathbf{U}$ is an orthogonal matrix containing the corresponding orthonormal eigenvectors of $\Ll$ as columns. The system of eigenvectors $\hat{G} = \{ u_1, \ldots, u_{n}\}$ forms an orthonormal basis of the signal space $\mathcal{L}(G)$. Given the Fourier basis $\hat{G}$, the graph Fourier transform $\hat{x}$ of a signal $x$, and the respective inverse Fourier transform are defined as
\begin{equation*}
\hat{x} := \mathbf{U^\intercal}x  = (u_1^\intercal x, \ldots, u_n^\intercal x)^{\intercal}, \quad \text{and} \quad x = \mathbf{U}\hat{x}.
\end{equation*}
In analogy to classical Fourier analysis, the basis elements $u_k$, $k \in \{1, \ldots, n\}$, can be regarded as Fourier modes of the graph and the entries $\hat{x}_k = u_k^\intercal x$ describe exactly the frequency content of the signal $x$ with respect to the Fourier mode $u_k$. 

As a general graph contains no inherent group structure, it is not possible to define the translation of a graph signal directly on the graph. The graph Fourier transform allows to bypass this problem and to introduce a generalized form of translation on graphs. This is done in terms of a convolution operator $\mathbf{C}_y$ acting on a signal $x$ as   
\begin{equation} 
\mathbf{C}_y x = \mathbf{U}\mathbf{M}_{\hat{y}}\mathbf{U^\intercal} x. 
\end{equation} 
Then, the convolution $\mathbf{C}_{\delta_{\node{v}}} x$ of a signal $x$ with the unit signal $\delta_{\node{v}}$ can be interpreted as a generalized translate of the signal $x$ on the graph $G$ by the node $\node{v}$. Indeed, if $G$ is endowed with a group structure and the Fourier basis $\hat{G}$ corresponds to the classical set of characters of the group $G$, then $\mathbf{C}_{\delta_{\node{v}}} x = x(\cdot - \node{v})$ is precisely a shift of the signal $x$ by the group element $\node{v}$. 
More information about different applications of the graph Fourier transform, for instance, as analysis, filtering or decomposition tool, can, for instance, be found in \cite{erb2019,Ortega2018,shuman2016,StankovicDakovicSejdic2019}. 

\subsection{Positive definite Graph Basis Functions (GBFs)} 
\label{sec:GBF}
Graph basis functions (GBFs) are simple and efficient tools for kernel-based interpolation and approximation methods on graphs \cite{erb2019b,erb2020}. 
In this GBF-method, the approximation spaces are built upon generalized shifts of a GBF $f$ such that every signal $x$ is approximated by a linear combination
\[x_*(\node{v}) = \sum_{i=1}^N c_i \mathbf{C}_{\delta_{\node{w}_i}} f(\node{v})\]
of generalized shifts $\mathbf{C}_{\delta_{\node{w}_i}} f$. The coefficients $c_i$ are calculated based on the information of $x$ on a sampling set $W = \{\node{w}_1, \ldots, \node{w}_N\} \subset V$. In this sense, the idea of GBFs is closely related to well-known theories in the literature on approximation with radial basis functions (RBFs)in the Euclidean space \cite{SchabackWendland2003,We05}, or spherical basis functions on the unit sphere.

An important prerequisite for the uniqueness of the approximant $x_*$ is the positive definiteness of the GBF $f$. A signal $f \in \mathcal{L}(G)$ is called positive definite (see \cite{erb2019}) if the kernel matrix 
\[ \mathbf{K}_{f} = \begin{pmatrix} \mathbf{C}_{\delta_{\node{v}_1}} f(\node{v}_1) & \mathbf{C}_{\delta_{\node{v}_2}} f(\node{v}_1) & \ldots & \mathbf{C}_{\delta_{\node{v}_n}} f(\node{v}_1) \\
\mathbf{C}_{\delta_{\node{v}_1}} f(\node{v}_2) & \mathbf{C}_{\delta_{\node{v}_2}} f(\node{v}_2) & \ldots & \mathbf{C}_{\delta_{\node{v}_n}} f(\node{v}_2) \\
\vdots & \vdots & \ddots & \vdots \\
\mathbf{C}_{\delta_{\node{v}_1}} f(\node{v}_n) & \mathbf{C}_{\delta_{\node{v}_2}} f(\node{v}_n) & \ldots & \mathbf{C}_{\delta_{\node{v}_n}} f(\node{v}_n)
\end{pmatrix} \]
is positive definite.
It was shown in \cite{erb2019b} that a signal $f$ is positive definite if and only if $\hat{f}_k > 0$ for all $k \in \{1, \ldots, n\}$. Further, the kernel function $K_f$ given by $K_f(\node{v},\node{w}) := \mathbf{C}_{\delta_{\node{w}}}f(\node{v})$ has the Mercer decomposition
\[ K_f(\node{v},\node{w}) = \mathbf{C}_{\delta_{\node{w}}}f(\node{v}) = \sum_{k=1}^n \hat{f}_k \, u_k(\node{v}) \, u_k(\node{w}).\]
In this way, a positive definite GBF $f$ induces, in a natural way, an inner product $\langle x,y \rangle_{K_f}$ and a norm $\| x \|_{K_f}$ by
\[ \langle x , y \rangle_{K_f} = 
\sum_{k=1}^n \frac{\hat{x}_k \, \hat{y}_k}{\hat{f}_k} = \hat{y}^\intercal \mathbf{M}_{1/\hat{f}} \, \hat{x} \quad \text{and} \quad \| x \|_{K_f} = \sqrt{\sum_{k=1}^n \frac{\hat{x}_k^2}{\hat{f}_k}}. \]
The space $\mathcal{L}(G)$ of signals endowed with this inner product is a reproducing kernel Hilbert space $\mathcal{N}_{K_f}$ with the reproducing kernel given as $K_f$.

\subsection{Stationary Gaussian Processes on Graphs}
We assume now that $X = ( X(\node{v}_1), \ldots, X(\node{v}_n))$ is a non-degenerate Gaussian random process on the graph $G$, i.e. that $X$ is a $n$-variate Gaussian random variable with a positive definite covariance matrix. We further assume that $X$ is stationary in the following sense:

\begin{dfn}
A Gaussian random process $X$ on the graph $G$ is called (wide-sense) stationary if the following two conditions are satisfied \cite{Perraudin2017}:
\begin{enumerate}[label = (\roman*)]
\item The expectation value $\Exp (X(\node{v}_i)) = \mu$ is constant for all nodes $\node{v}_i$ of $G$.
\item The covariance matrix of the Gaussian process is invariant under the generalized shift operator $\mathbf{C}_{\delta_{\node{v}_i}}$ and given by the positive definite matrix
\[\Exp \left((X(\node{v}_i) - \mu)(X(\node{v}_j) - \mu)\right) = \mathbf{K}_{f,\subind{i}{j}} = 
\mathbf{C}_{\delta_{\node{v}_i}} f(\node{v}_j).\]
\end{enumerate}
\end{dfn} 
In particular, every stationary Gaussian random process $X$ is uniquely determined by the mean value $\mu \in \Rr$ and a positive definite GBF $f$ providing the covariance matrix $\mathbf{K}_{f}$ of the process. Further, the density function $\varphi_X$ of the stationary Gaussian process $X$ is given by
\[ \varphi_X(x) = \frac{\exp(-\frac12 
(x-\mu \1 )^\intercal \mathbf{K}_f^{-1}(x-\mu \1 ))}{\sqrt{(2\pi)^n |\mathbf{K}_f|}},\]
where $\1 = (1, \ldots, 1)^\intercal \in \Rr^n$ and $|\mathbf{K}_f|$ denotes the determinant of $\mathbf{K}_f$.

\subsection{Gaussian process regression (kriging)} \label{sec:kriging}

Gaussian processes provide simple stochastic models for data regression. We consider additive white noise $\epsilon$ on the graph $G$ as a stationary Gaussian process with zero mean and covariance matrix $\sigma^2 \mathbf{I}_n$. Then, if $X$ denotes a stationary Gaussian process on $G$ with mean $\mu = 0$ and covariance matrix $\mathbf{K}_f$ the random signal
\begin{equation} \label{eq:Gaussianregression} Y = X + \epsilon 
\end{equation}
is also a stationary Gaussian process with mean $\mu = 0$ and covariance matrix $\mathbf{K}_f + \sigma^2 \mathbf{I}_n$. From $N$ known samples $Y(\node{w}_i) = y_i$, $i \in \{1, \ldots, N\}$, on a given sampling set $W = \{\node{w}_1, \ldots, \node{w}_N\}$ our goal is to find a predictor $x_*$ of the process $X$ based on the given sampling information and the model in \eqref{eq:Gaussianregression}. As $X$ and $Y$ are both Gaussian, the $N+1$ dimensional random vector $(Y(W), X(\node{v})) = (Y(\node{w}_1), \ldots, Y(\node{w}_N), X(\node{v}))$ is also Gaussian for any node $\node{v} \in V$. The corresponding normal distribution is given as
\[ \begin{pmatrix} Y(W) \\ X(\node{v}) \end{pmatrix} \sim 
N\left( \vect{0}, \begin{pmatrix} \mathbf{K}_{f,W} + \sigma^2 \mathbf{I}_n &\;\;  \mathbf{C}_{\delta_{W}} f(\node{v})\\ \mathbf{C}_{\delta_{W}} f(\node{v}) ^{\intercal} & \mathbf{K}_{f,\node{v}} \end{pmatrix} \right),\]
in which the covariance matrix is built upon $\mathbf{K}_{f,\node{v}} = \mathbf{C}_{\delta_{\node{v}}}f(\node{v})$, the positive definite matrix $\mathbf{K}_{f,W} \in \Rr^{N \times N}$, and the vector $\mathbf{C}_{\delta_{W}} f(\node{v}) \in \Rr^{N}$ given as
\begin{equation} \label{eq:partsofcovariancematrix} 
 \mathbf{K}_{f,W} = \begin{pmatrix} \mathbf{C}_{\delta_{\node{w}_1}}f(\node{w}_1) & \ldots & \mathbf{C}_{\delta_{\node{w}_N}}f(\node{w}_1) \\
\vdots & \ddots & \vdots \\
\mathbf{C}_{\delta_{\node{w}_N}}f(\node{w}_1) & \ldots & \mathbf{C}_{\delta_{\node{w}_N}}f(\node{w}_N)
\end{pmatrix}, \quad \mathbf{C}_{\delta_{W}} f(\node{v}) = \begin{pmatrix} \mathbf{C}_{\delta_{\node{w}_1}}f(\node{v})  \\
\vdots \\
\mathbf{C}_{\delta_{\node{w}_N}}f(\node{v}) 
\end{pmatrix}.
\end{equation}
For this, also the conditional distribution of the random variable $X(\node{v})$ given the information $Y(\node{w}_i) = y_i$, $i \in \{1, \ldots, N\}$ is a Gaussian distribution with the (conditional) expectation value
\begin{equation} \label{eq:representertheorem}
x_{*}(\node{v}) = \Exp(X(\node{v})| \ Y(\node{w}_1) = y_1, \ldots, Y(\node{w}_N) = y_N ) =  \sum_{i = 1}^N c_i \mathbf{C}_{\delta_{\node{w}_i}}f(\node{v}), 
\end{equation}
and the coefficients $(c_1, \ldots, c_N)^\intercal$ calculated as (see \cite{Wahba1990}) 
\begin{equation} \label{eq:computationcoefficients} 
 \begin{pmatrix} \node{c}_1 \\ \vdots \\ c_N \end{pmatrix}
= \left( \mathbf{K}_{f,W} + \sigma^2 \mathbf{I}_N \right)^{-1}\begin{pmatrix} y_1 \\ \vdots \\ y_N \end{pmatrix}.
\end{equation}
Furthermore, the standard deviation of this conditional Gaussian distribution is given as 
\begin{equation} \label{eq:gaussianprocesspowerfunction} \Power_W(\node{v}) = \Big( \mathbf{K}_{f,\node{v}} -  \mathbf{C}_{\delta_{W}} f(\node{v})^{\intercal} (\mathbf{K}_{f,W} + \sigma^2 \mathbf{I}_n )^{-1} \mathbf{C}_{\delta_{W}} f(\node{v})\Big)^{1/2}.
\end{equation}
Note that this posterior standard deviation $\Power_W(\node{v})$, and, in the same way, the squared variance, depend on the node set $W$ but not on the sampling values $y_1, \ldots, y_N$. It is a measure of uncertainty for the process $X$ on a node $\node{v}$ if (arbitrary) information is provided on the sampling set $W$.

\section{A new model for influence maximization on graphs based on Gaussian process variance minimization}\label{sec:new_model}

Our model to determine a set $\WIM_N \subset V$ of $N$ graph nodes with a maximal influence on the graph $G$ is based on the solution of the following optimization problem:

\begin{equation}\label{imkernelbased}
\WIM_N=\mbox{arg}\min_{|W|=N} \underbrace{\max_{\node{v} \in V} | \Power_W(\node{v}) |}_{\Phi(W)}.
\end{equation}
Observe that $\Power_W$, and thus $\WIM_N$, has an hidden dependence on the noise variance $\sigma^2$ (see \eqref{eq:gaussianprocesspowerfunction}). Since our aim is to select influential nodes independently of a particular graph signal, we set from now on $\sigma:=0$.

\begin{ppt}
The model \eqref{imkernelbased} for IM has the following properties:
\begin{itemize}
    \item[$(i)$] The model function is bounded by
    \[0 \leq \Phi(W) \leq \max_{\node{v} \in V} \sqrt{\Kk_{f,\node{v}}} \quad \text{for all $W \subset V$}. \]
    \item[$(ii)$] The model function is monotonic (see \cite{Schaback1995}), i.e.,
    \[\Phi(\WIM_{N+1}) \leq \Phi(\WIM_N).\]
\end{itemize}
\end{ppt}
This model contains a rather large set of parameters that have to be determined a priori or extracted from given data. The main components are
\begin{itemize}
    \item[$\bullet$] In the construction of the graph $G$ from a set of given nodes and edges, the Laplacian $\Ll$ is not always a priori fixed. The Laplacian $\Ll$ contains all the connection weights representing the local influence of the nodes to their neighboring nodes. 
    \item[$\bullet$] Once the Laplacian $\Ll$ is determined, the choice of the kernel $\Kk_f$ determines the global spread of information on the graph $G$. Such a kernel can be given a priori or calculated from data by a learning approach. 
    With respect to the last point, an approach to kernel selection based on the training and inference via sampling random Fourier features has been proposed for kernels on $\Rr^d$ in \cite{li2019}, and a variational improvement is given in \cite{Zhen2020}. Additionally, methods based on the maximization of a likelihood function are very popular in machine learning (see e.g. \cite{Fasshauer2015}, Chapter 14).
    Two popular kernels for a priori choices are given in Example \ref{exa:GBFs}.  
\end{itemize}

\begin{exa} \label{exa:GBFs}
Two well-known examples of positive definite GBF-kernels $\Kk_f$ on a graph $G$ are the following \cite{erb2019b}: 
\begin{enumerate}
\item[$(i)$] The diffusion kernel on a graph \cite{KondorLafferty2002} is given by 
\[\mathbf{K}_{f_{e^{-t \mathbf{L}}}} = e^{ -t \mathbf{L}} = \sum_{k=1}^n e^{ -t \lambda_k} u_k u_k^{\intercal},\]
where $\lambda_k$, $k \in \{1, \ldots, n\}$, denote the eigenvalues of the Laplacian $\Ll$. This matrix is positive definite for all $t \in \Rr$. The coefficients $e^{-t \lambda_k}$ in the given Mercer decomposition are precisely the Fourier coefficients
of $f_{e^{-t \mathbf{L}}}$. 

\item[$(ii)$] The variational spline
$$ \mathbf{K}_{f_{(\epsilon \mathbf{I}_n + \mathbf{L})^{-s}}} = (\epsilon \mathbf{I}_n + \mathbf{L})^{-s} = \sum_{k=1}^n \frac{1}{(\epsilon + \lambda_k)^s} u_k u_k^{\intercal}$$
is positive definite for $\epsilon > - \lambda_1$ and $s > 0$. These type of GBF-kernels
were investigated in \cite{Pesenson2009,Ward2018interpolating} in terms of signal interpolation. Interpolants based on the translates of this GBF minimize an energy functional and are referred to as variational or polyharmonic splines.  
\end{enumerate}
\end{exa}

When leaving the graph setting for a short moment, similar models as the one given in \eqref{imkernelbased} can be found in a multitude of applications related to geosciences or in approximation theory. This provides us with a better understanding on how the model can be interpreted in the graph setting.

\noindent \textbf{The model \eqref{imkernelbased} in geostatistical kriging.} In geostatistics, methods related to the Gaussian process regression described in Section \ref{sec:kriging} are usually referred to as kriging methods. In kriging, the square of the Power function $\Power_W$ is called the kriging variance and its global minimization is used for the quantitative design of infill sampling strategies in the investigation of geological sites. In particular, optimization problems as the one in \eqref{imkernelbased} are widely used criteria for the location of new sampling positions \cite{bras-rodriguez-iturbe,ScheckChou1983}. 

\noindent \textbf{The model \eqref{imkernelbased} in approximation theory:}
The power function plays a central role in the error estimation of kernel-based interpolation in approximation theory. The notion has first been introduced in \cite{Schaback1995}, and later used in a multitude of works. We refer to Chapter 11 in \cite{We05} for a comprehensive treatment of this topic. Minimization criteria as in \eqref{imkernelbased} in the context of approximation theory are discussed in the next section.

\section{P-greedy for variance minimization}\label{sec:p_greedy}
The determination of the set $\WIM_N \subset V$ from the exact solution of \eqref{imkernelbased} requires to consider all the $\binom{n}{N} \geq \left(n/N\right)^N$ possible subsets of $N$ elements among the $n$ nodes of the graph. This combinatorial problem is practically intractable already for moderate-sized graphs.

To overcome this limitation, we replace the exact criterion \eqref{imkernelbased} with an approximation based on a very simple, yet effective greedy procedure. Namely, starting from the empty set of nodes $\WIM_0=\emptyset$, at each iteration $k=1, \dots, N$ we update it by adding a new node as
\begin{align}\label{eq:p_greedy}
&\node{w}_k\in \argmax\limits_{\node{w}\in V \setminus \WIM_{k-1}} {\Power_{\WIM_{k-1}}(w)},\\
&\WIM_{k}:=\WIM_{k-1}\cup \{\node{w}_{k}\}\nonumber,
\end{align}
where the inclusion relation ``$\in$'' denotes the fact that the node realizing the maximum may be non-unique. Moreover, \eqref{eq:gaussianprocesspowerfunction} implies that $\Power_{\WIM_0}(w)= \sqrt{\mathbf{K}_{f,\node{w}}}$ for all $\node{w} \in V$, and thus the selection is well defined for all $k$.
Finally, observe that the method can also be run to enrich an initial set of nodes, i.e., one may start with $\WIM_0=W_0$, with any $W_0\subset V$.

This method has been introduced in \cite{DeMarchi2005} under the name of $P$-greedy algorithm in the context of kernel approximation on $\Rr^d$. Despite its simplicity, the method has been proven to be asymptotically optimal (see  \cite{Santin2017,Wenzel2021}), i.e., greedily selected nodes provide the same rate of decay of $\Power_{\WIM_{N}}$ as optimally chosen ones. 
Moreover, the algorithm has a very efficient implementation (see \cite{Pazouki2011,Santin2021}). Indeed, an efficient formula is available to update $\Power_{\WIM_{k-1}}$ to $\Power_{\WIM_{k}}$, and thus \eqref{eq:p_greedy} requires only the update of the solution of a $k$-dimensional triangular linear system and a simple linear search over $V \setminus \WIM_{k-1}$ at each iteration.

Observe that the theoretical properties hold in principle also in the present graph setting, but they are not particularly relevant as they are tailored to the asymptotic case $N\to\infty$, which makes no sense in this setting. Nevertheless, the excellent performances of the method that will be observed in Section \ref{sec:numerics} seem to suggest that similar quasi-optimality may be proven here. We leave the investigation of this point open for future study.

On the other hand, the computational aspects remain the same and in particular the same implementation of the method is seamlessly extended to our setting, resulting in a very efficient algorithm.

We remark that the $P$-greedy algorithm has been extended to the case of matrix-valued kernels in \cite{Wittwar2019}. In our setting, this would allow to approximate vector-valued signals over the graph.

\section{Numerical experiments}\label{sec:numerics}
In this section we test our method under different points of view. In Section \ref{sec:exp1} we set some parameters and configurations to clarify the setup of the following experiments. Section \ref{sec:exp2} is devoted to an in-depth data-driven tuning of the algorithm in order to obtain an optimized method, which is then compared with state of the art alternatives in Section \ref{sec:exp3}.

\subsection{Experimental setup}\label{sec:exp1}
All the numerical tests are realized with a Python implementation of the method discussed in the previous sections, which builds on the Python library NetworkX \cite{Hagberg2008} for graph representation. The implementation, including the code to replicate the experiments, is made public at \cite{GBF_code}. 

We use as test cases some small- to medium-sized graphs that are described in Table~\ref{tab:graphs}, and that can be generated by the code in \cite{GBF_code}. For visualization convenience only, all these graphs have a two-dimensional position attribute associated with each node. 

\begin{table}[htbp] \caption{Graphs used in the numerical experiments}
\label{tab:graphs} 
\begin{center}
\begin{tabular}{|p{12mm}|p{12mm}|p{12mm}|p{60mm}|}
\hline 
ID  & N. nodes & N. edges& Description\\
\hline 
\texttt{bunny} & $1035$ & $9468$ & The nodes are obtained by projecting into the plane the points of the {\em Stanford bunny} \cite{Curless1996}, and thinning the points with a separation radius of $0.0025$. The nodes that are within a radius of $0.01$ are then linked.\\
&&&\\
\texttt{sensor1} & $79$ & $210$ & This sensor graph consisting of $79$ randomly chosen points in the plane has been generated for this work. \\
\hline
\end{tabular}
\end{center}
\end{table}

The influence maximization algorithm of Section \ref{sec:p_greedy} is run on the entire set of nodes (i.e., each node may potentially be chosen). The constant signal with value $1$ on all nodes is used to validate the algorithm, namely residuals are computed with respect to the approximation of this signal. The algorithm is terminated when a maximal number of nodes are selected, or when either the maximal value of the squared standard deviation, or the maximal valued of the residual fall below the tolerance $\tau:=10^{-12}$. This ensures a certain stability in the algorithm (see \cite{Santin2021}).
As GBFs we use the diffusion and the variational spline kernels (see Example \ref{exa:GBFs}), with parameters that are defined in each experiment.

\begin{figure}[htbp]
	\centering
    \begin{tabular}{cc}
	\includegraphics[width=.45\textwidth]{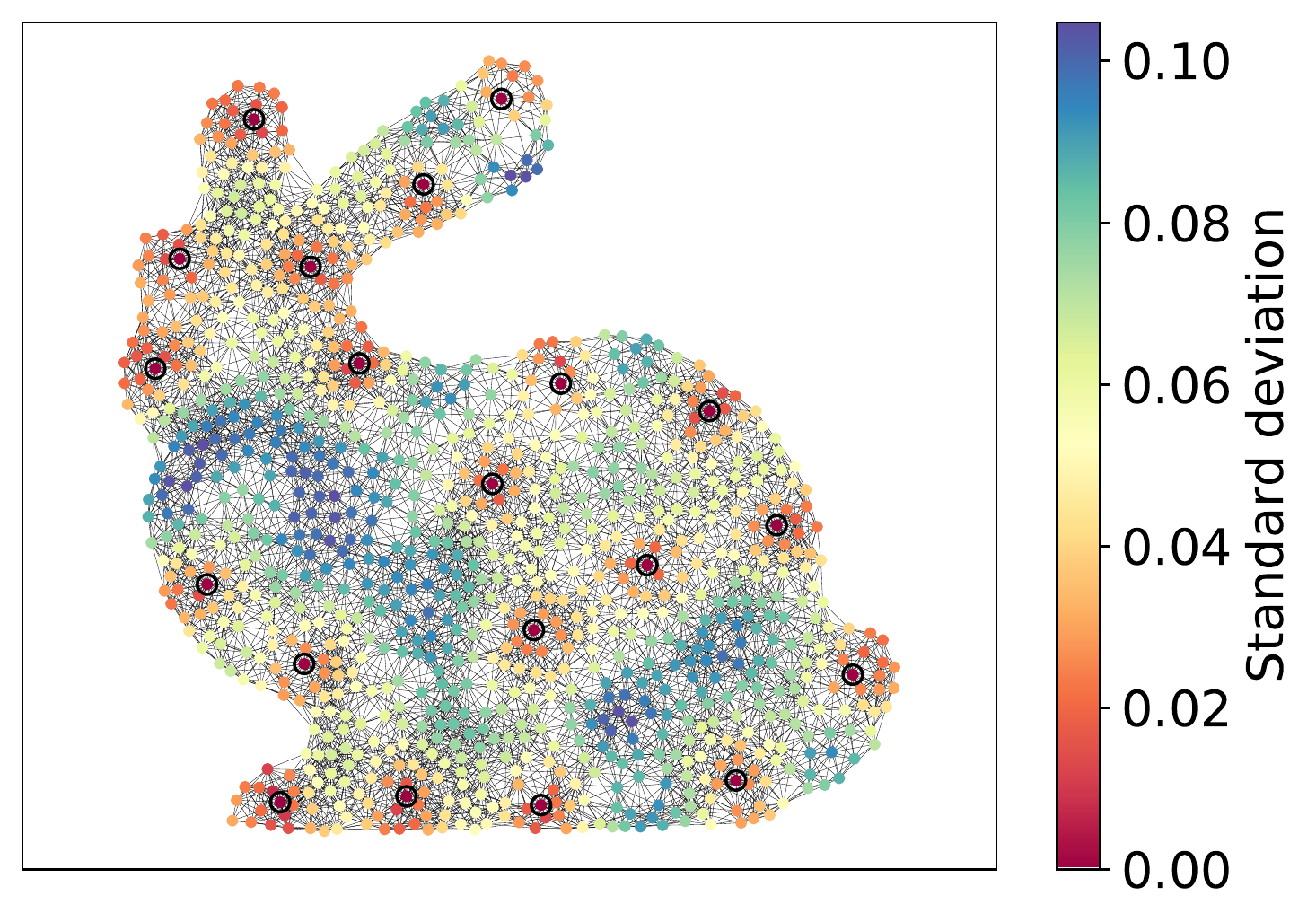}
	\includegraphics[width=.45\textwidth]{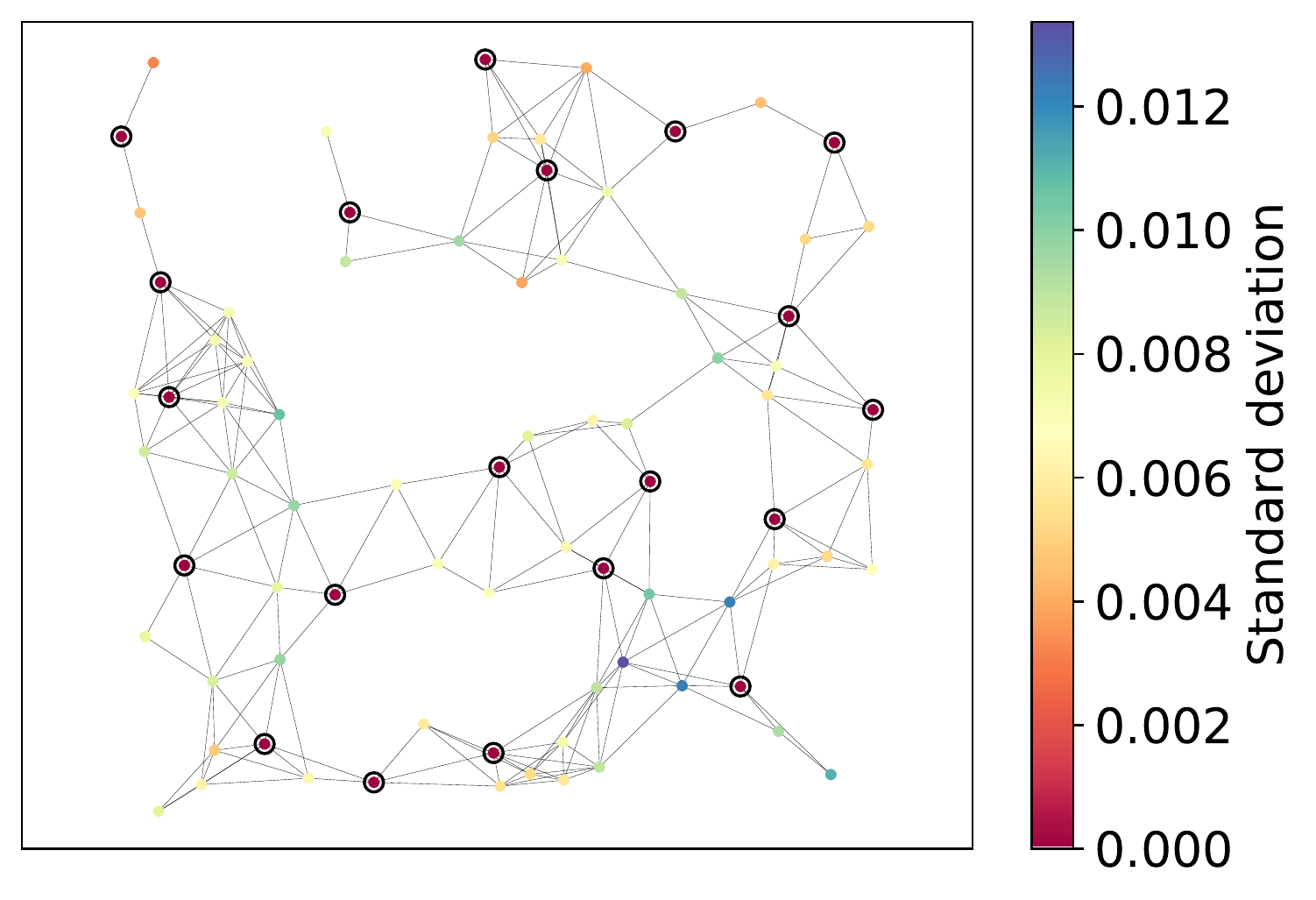}
	\end{tabular}
	
	\caption{Example of the first $20$ points (circled nodes) selected by the new algorithm using a diffusion kernel on \texttt{bunny} (left) and \texttt{sensor1} (right). The nodes are colored according to the value of the standard deviation $\Power_{W_{20}^{(\mathrm{IM})}}$.}
	\label{fig:exp1}
\end{figure}

As an example, we visualize in Figure~\ref{fig:exp1} the first $20$ nodes selected by the algorithm on \texttt{bunny} (left panel) and \texttt{sensor1} (right panel), using a diffusion kernel with parameter $t=-10$. 

\begin{rem}
It is worth noticing that in the first case the nodes appear to be spatially uniform, as one may expect since the graph has uniformly distributed nodes, and this is in line with the behavior of the node selection algorithm when applied in $\mathbb R^d$ (see \cite{Wenzel2021}).
On the other hand, also in the second case a kind of uniformity seems to be present, even if constrained to the graph's topology. The investigation of the properties of these node distributions is an interesting topic that we intend to address in the future.
\end{rem}

\subsection{Data-driven model learning}\label{sec:exp2}
Since our definition of influence is depending on the chosen kernel (including its parameters), it is of upmost relevance to devise a systematic methodology to learn the kernel from data.

As an initial attempt in this direction, we use here a cross validation approach. Namely, we use standard $k$-fold cross validation to optimize the kernel and its parameters in order to minimize the residual in the approximation of the signal with constant value $1$. This target signal is chosen as it equally weights all nodes in the graph, and thus its approximation seems to be a suitable proxy to model a uniform spread of the information. 

To be more specific, on \texttt{bunny} we compare a variational spline GBF with parameters $\epsilon=0.01$, $s=-1$, and a diffusion GBF with $t=-10$, with their optimized variants. The optimization is realized with $5$-fold cross validation over the parameter grids described in Table \ref{tab:cross_val}. As a model selector we use the mean error in the approximation of the constant signal over the entire graph. The selected parameters are reported also in Table \ref{tab:cross_val}. We stress that it would be easily possible to include the kernel itself as an optimizable parameter, but here we prefer to derive two different models, one per kernel, in order to compare them.

\begin{table}[htbp] \caption{Range of the search grid for the parameter optimization (first row) and selected parameters (second row) for the variational spline and the diffusion GBFs. Each parameter range is discretized into $25$ logarithmically equally spaced values.}
\label{tab:cross_val} 
\begin{center}
\begin{tabular}{|c|cc|c|}
\hline
&\multicolumn{2}{c|}{variational spline} & diffusion\\ 
 & $\epsilon$ & $s$ &  $t$ \\
\hline
search interval  & $[10^{-16}, 10^0]$ &$[-10^{1}, -10^{-1}]$ & $[-10^{2}, -10^{-2}]$\\
optimized &$10^{-6}$& $-2.15$ & $-31.62$\\
\hline
\end{tabular}
\end{center}
\end{table}

\begin{figure}[htbp]
	\centering
    \includegraphics[width=.9\textwidth]{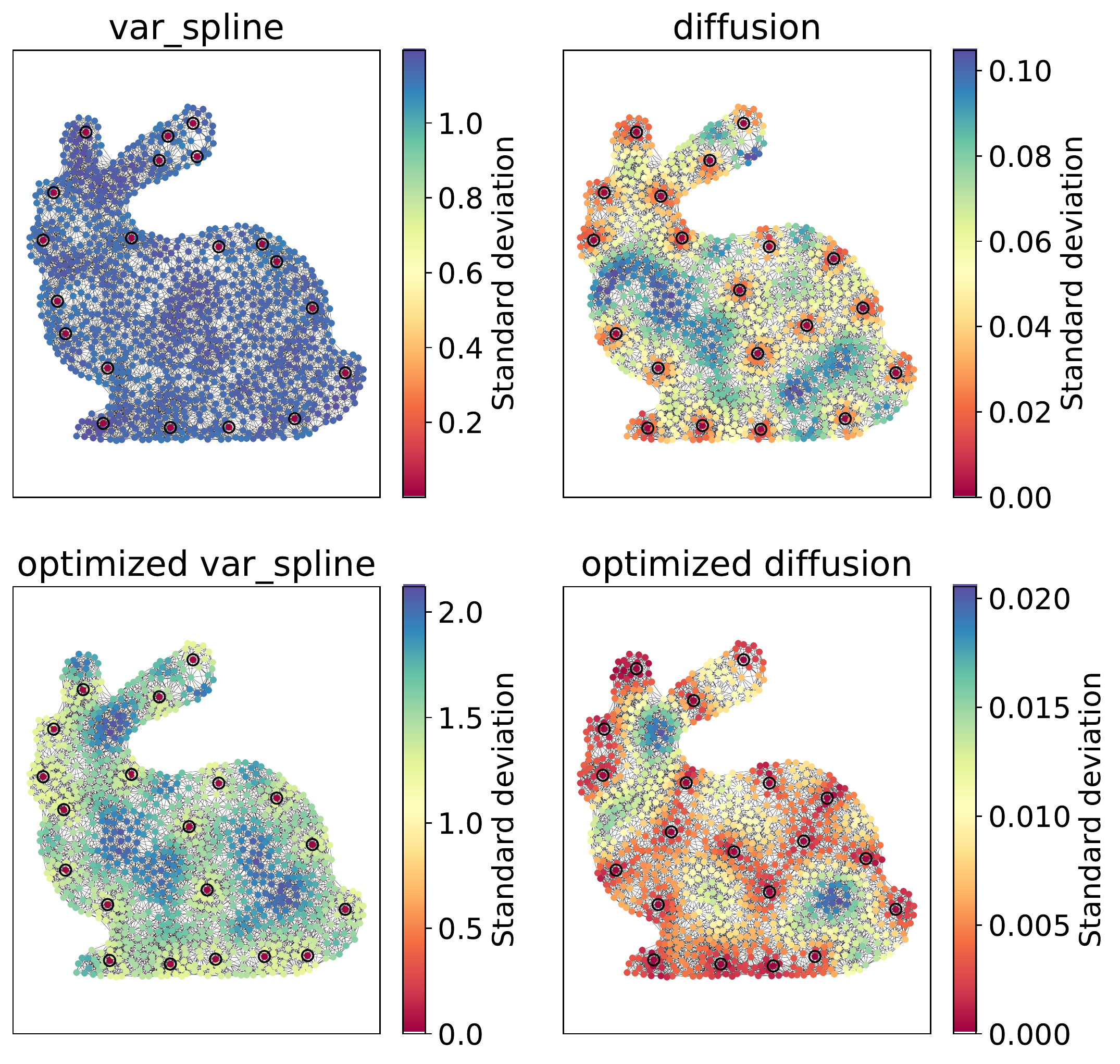}
	\caption{First $20$ nodes selected by the new algorithm on \texttt{bunny} using the variational spline GBF (first column) or the diffusion GBF (second column), before (first row) and after the optimization of their parameters (second row).} 
	\label{fig:exp2_points}
\end{figure}

\begin{figure}[htbp]
	\centering
\includegraphics[width=.9\textwidth]{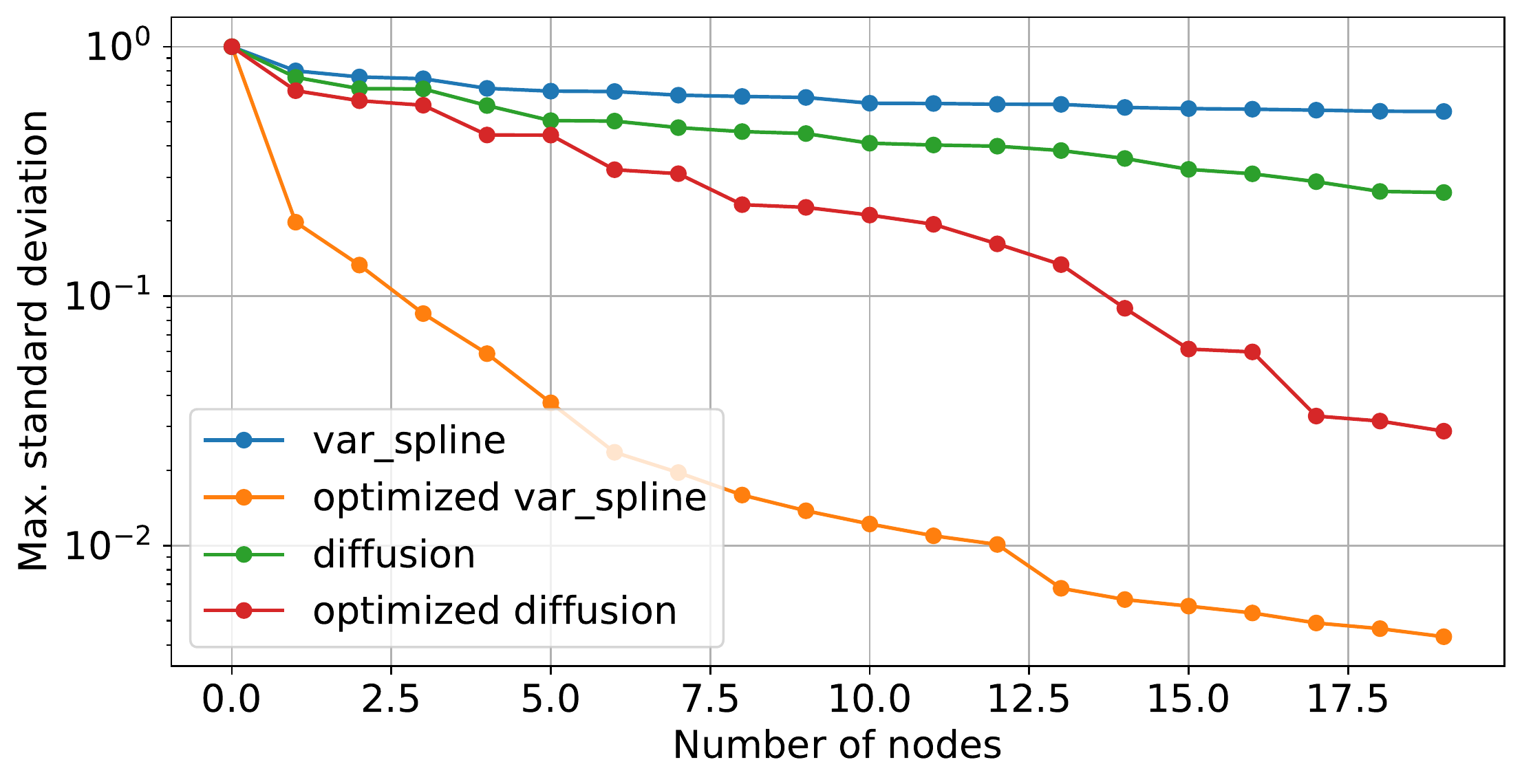}
		\caption{Decay of the standard deviation as a function of the number of the nodes selected by the new algorithm on \texttt{bunny} using the variational spline GBF or the diffusion GBF, before and after the optimization of their parameters. All standard deviations are normalized to an initial value of $1$.} 
	\label{fig:exp2_std}
\end{figure}

The first $20$ points resulting from the four selections are depicted in Figure \ref{fig:exp2_points} and colored with the absolute value of the final standard deviation. Observe that this quantity is kernel-dependent, and thus different scales are present in the plots. It is immediately clear, especially for the variational spline, that optimized parameters yield better distributed nodes and thus possibly a more credible notion of influence maximization. This is even more evident and quantitatively assessed in Figure \ref{fig:exp2_std}, where we report the decay of the maximal standard deviation on the entire graph as a function of the selected nodes. In this case, to facilitate the comparison of the four configurations, we show standard deviations normalized by their initial maximum. The decay of the maxima is much faster for the optimized kernels, with significantly smaller values for the variational spline. 

Table \ref{tab:exp2_point_order} additionally reports the ids of the first selected nodes. Observe that there is a certain stability for the different configurations, and this is a good indication of the quality of the selection process. On the other hand, notice that slight changes in the order have already an effect in the decay of the standard deviation (see Figure \ref{fig:exp2_std}), and this points to the importance of a proper kernel selection to maximize the performances of the method.

\begin{table}[htbp] \caption{First $5$ nodes selected by the new algorithm on \texttt{bunny} using the variational spline GBF or the diffusion GBF, either optimized or not.}
\label{tab:exp2_point_order} 
\begin{center}
\begin{tabular}{|cc|cc|}
\hline
\multicolumn{2}{|c|}{variational spline} & \multicolumn{2}{c|}{diffusion}\\ 
not optimized & optimized &  not optimized  &  optimized \\
\hline
          4 &                     4 &          4 &                    4 \\
        730 &                   344 &        730 &                  730 \\
        164 &                   164 &        164 &                  776 \\
        776 &                    17 &        775 &                  164 \\
        459 &                   919 &        121 &                  793 \\

\hline
\end{tabular}
\end{center}
\end{table}

\subsection{Comparison with state of the art methods}\label{sec:exp3}

We compare now on \texttt{sensor1} the quality of the nodes selected by our method to other state of the art techniques, namely Pagerank (PR) \cite{Page1999} and Independent Cascade (IC) \cite{DAngelo2016}. For the first algorithm we use the implementation provided by NetworkX, while for the second we use an own implementation that is made available in \cite{GBF_code}. 

The algorithm IC is run with a spread probability $p=0.2$ and with $500$ independent runs at each iteration, and furthermore it is applied in a greedy mode, i.e., the algorithm is run sequentially to select each time the next most influential node. 
For our method we use also in this case the variational spline GBF, with parameters optimized in the same way as in the previous section, resulting into $\epsilon=-2.15\cdot10^{-11}$, $s=-1$ (the parameters are different from the ones of the previous section because the underlying graph is different).

\begin{figure}[htbp]
	\centering
	\includegraphics[width=.9\textwidth]{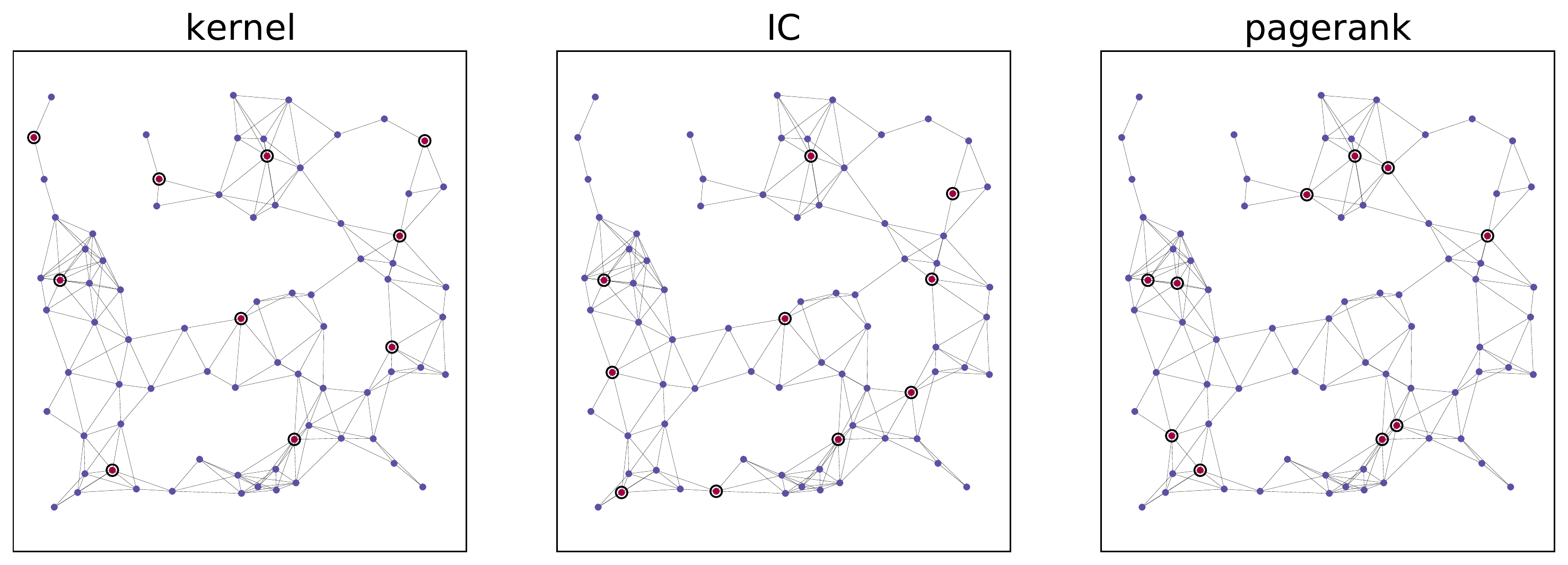}
	\caption{Distribution of the first $10$ nodes selected by the new algorithm (left), IC (center) and pagerank (right) when applied to \texttt{sensor1}.}
	\label{fig:exp3_points}
\end{figure}

The first $10$ nodes selected by each algorithm are shown in Figure~\ref{fig:exp3_points} and listed in Table~\ref{tab:exp3_point_order}, where we additionally report the $10$ nodes of higher degree. In this case, the three nodes $1, 23, 64$ are selected by all the three algorithms and are among the $10$ nodes of higher degree. Moreover, the new algorithm shares $5$ of the first $10$ nodes with PG, while IC and PG share $3$.

\begin{figure}[htbp]
	\centering
    \begin{tabular}{cc}
	\includegraphics[width=.45\textwidth]{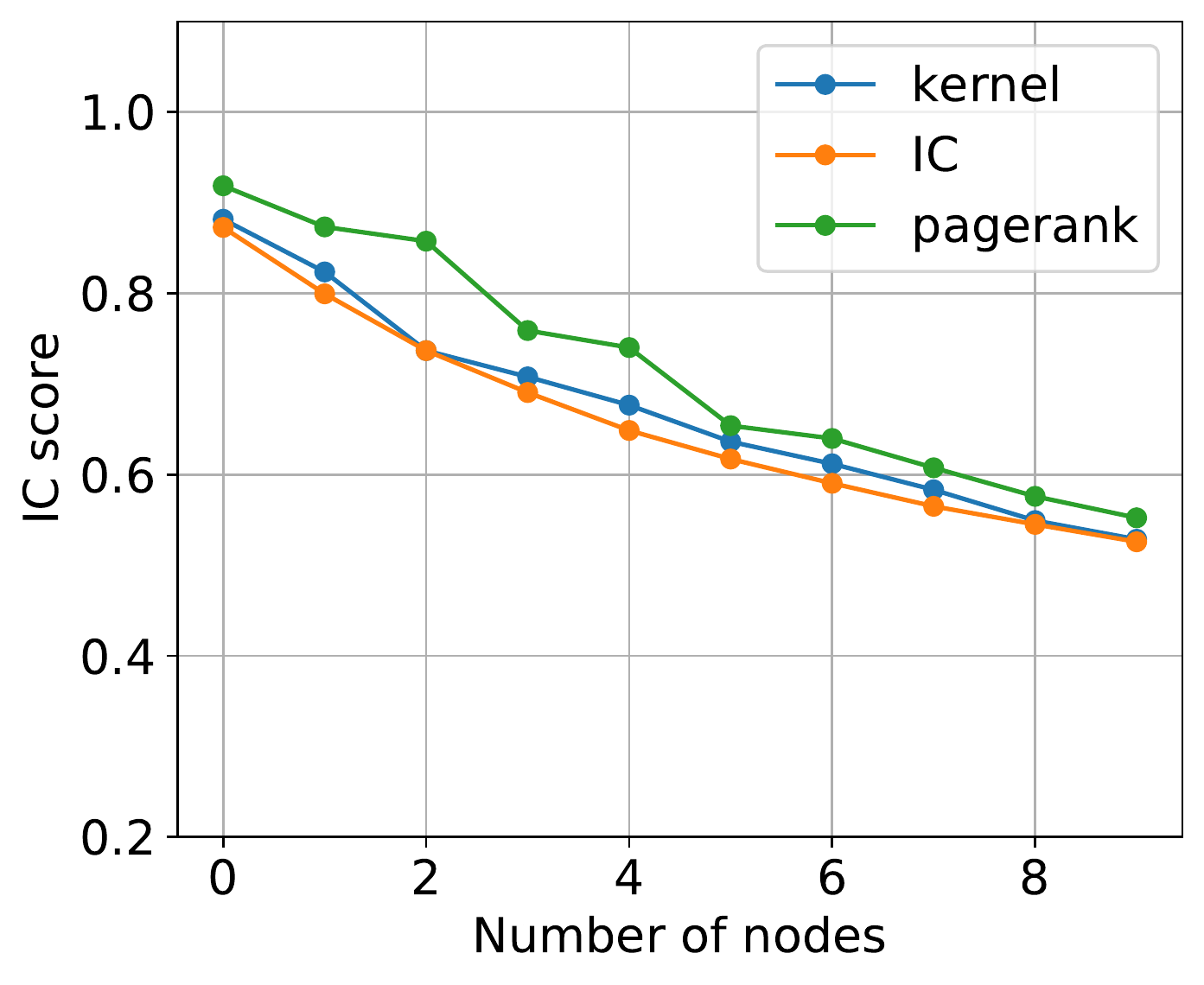}
	\includegraphics[width=.45\textwidth]{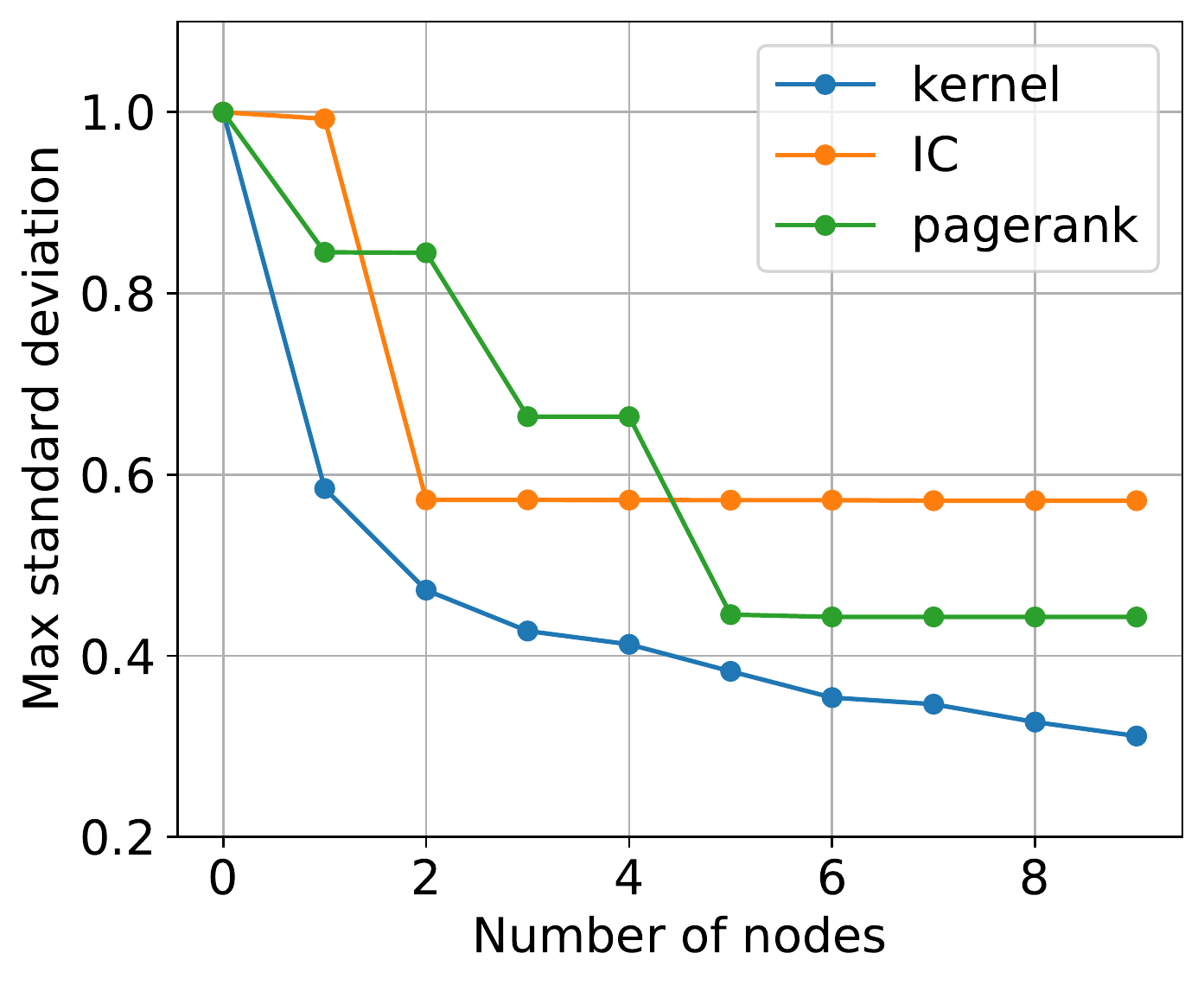}
\\
	\includegraphics[width=.45\textwidth]{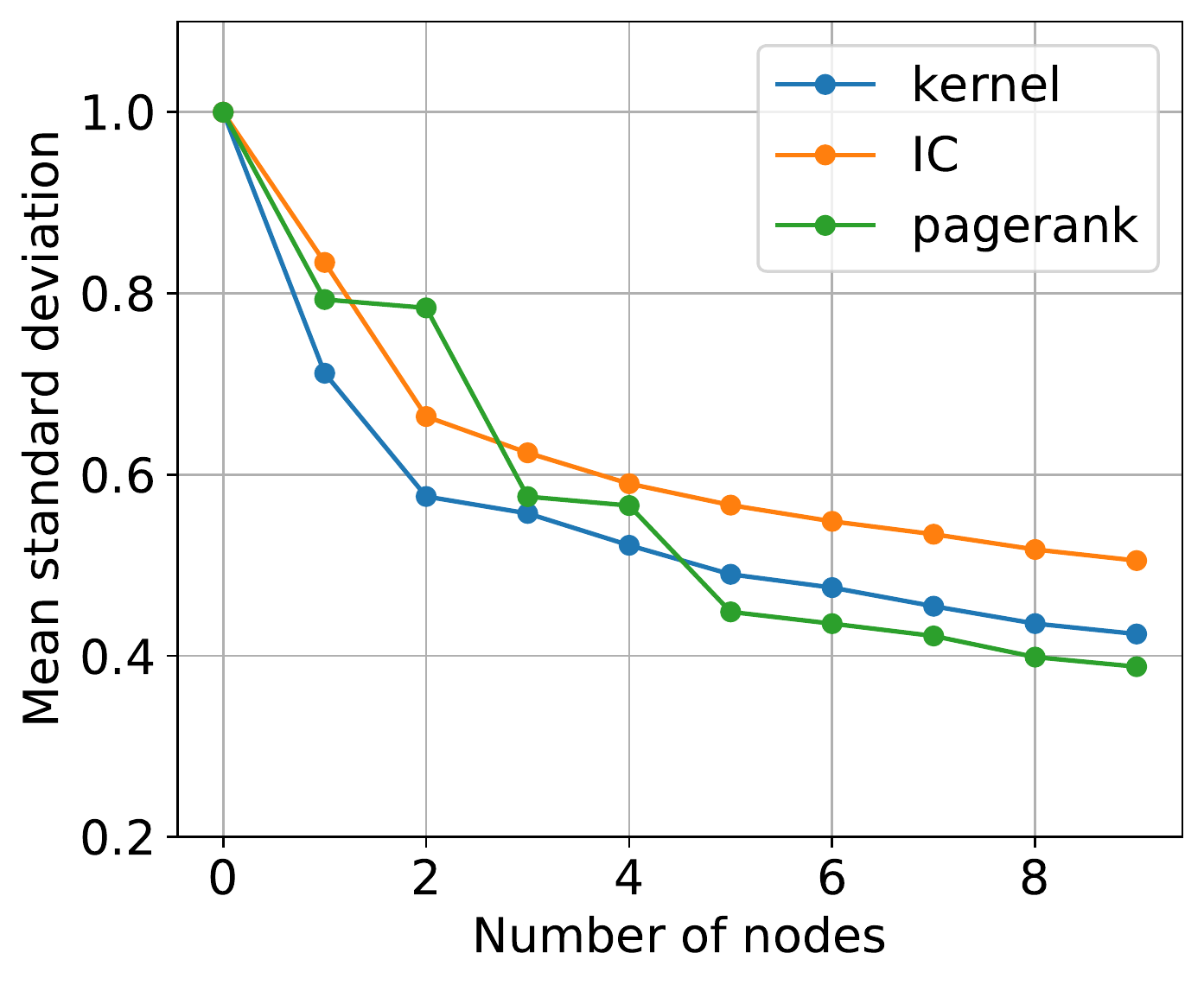}
	\end{tabular}
	\caption{Decay of the IC score (first line left), of the max standard deviation (first line right), of the mean standard deviation (second line) as a function of the number of nodes selected by the new algorithm, IC, and pagerank when applied to \texttt{sensor1}.}
	\label{fig:exp3_decays}
\end{figure}

To quantify the quality of the selected nodes, we compare them according to the two metrics defined by IC and by our algorithm.
Namely, for each incremental set of nodes we compute
the max value of the standard deviation over the entire graph using the optimized variational spline GBF, and the IC score, i.e., the fraction of nodes not reached by IC starting from the set of given nodes.
The decay of these metrics are shown in the first row of Figure~\ref{fig:exp3_decays}. As expected, both the current method and IC minimize the respective score, but it is interesting to observe that the new method performs almost optimally also with respect to the IC metric. If one instead considers the mean standard deviation (second row of Figure~\ref{fig:exp3_decays}), it is remarkable that PG outperforms both methods, even if slightly.

In particular, since our method can be executed in a fraction of the time required by IC, this is an indication that the new influence maximization algorithm can be used as a competitive alternative to state of the art methods.

\begin{table}[htbp] \caption{First $10$ nodes selected by the new algorithm, IC, and pagerank when applied to \texttt{sensor1}, and $10$ nodes of highest 
degree. The nodes that are selected by all algorithms are in boldface fonts, while the ones selected both by PR and by the new algorithm are underlined.}
\label{tab:exp3_point_order} 
\begin{center}
\begin{tabular}{|c|cccccccccc|}
\hline
kernel & \underline{\bf{64}} &  \underline{\bf{1}} &  \underline{\bf{23}} &   8 &   \underline{6} &  \underline{43} &  74 &  16 &  65 &  73 \\
IC &  \bf{23} &  \bf{64} & \bf{1} &   0 &  58 &  16 &  14 &  22 &  17 &  76 \\
PR &  \underline{\bf{1}} &   \underline{6} &  56 &  \underline{\bf{23}} &  51 &  \underline{\bf{64}} &  71 &  24 &  30 &  \underline{43} \\
node degree &  \bf{23} &  \bf{64} &  71 &  \bf{1} &  24 &  35 &  42 &  53 &  60 &   6 \\
\hline
\end{tabular}
\end{center}
\end{table}

\section{Conclusion}
In the context of mathematical foundations of DS, we presented a novel method for influence maximization on graphs. The proposed deterministic approach is based on GBF-kernels and the minimization of a variance term related to Gaussian process regression on graphs. In order to select the influential nodes of a graph, we implemented a cost-efficient approximate minimization of the variance by an efficient P-greedy algorithm. We could experimentally show that our scheme is more efficient than respective greedy schemes involving stochastic spread models, for instance, the independent cascade model. An emerging issue for our model is the proper selection of the covariance kernel which has a direct impact on the performance of the method. To overcome this problem, we tested a data-driven approach to determine the kernel and used machine learning techniques for the parameter tuning of the model. Challenging questions for future research include smart strategies for a data-driven extraction of the kernel,  machine learning methodologies for an optimal selection of the model parameters, as well as theoretical questions on possible distributions of the selected nodes.

\section*{Acknowledgments}
This work was partially supported by INdAM-GNCS. This research has been accomplished within the two Italian research groups on approximation theory RITA and UMI T.A.A.

\bibliography{biblio.bib}
\bibliographystyle{abbrv}

\end{document}